\documentclass[runningheads]{llncs}

\usepackage[final,year=2026]{eccv}

\usepackage{eccvabbrv}

\usepackage{graphicx}
\usepackage{booktabs}

\usepackage[accsupp]{axessibility}  

\usepackage{hyperref}

\usepackage{booktabs}
\usepackage{multirow}
\usepackage{graphicx}
\usepackage{tabularx}

\begin{document}

\title{D²Turb:Depth-Aware Simulation and Decoupled Learning for Single-Frame Atmospheric Turbulence Mitigation} 

\titlerunning{Abbreviated paper title}

\author{Zixiao Hu\inst{1} \and
Tianyu Li\inst{1} \and
Guoqing Wang\inst{1} \and
Wei Li\inst{2} \and
Guoguo Xin\inst{3} \and
Xun Liu\inst{2} \and
Peng Wang\inst{1}}

\authorrunning{Z.~Hu et al.}

\institute{University of Electronic Science and Technology of China \and
Beijing Institute of Space Mechanics and Electricity \and
School of Physics, Northwest University Xi'an}

\maketitle

\begin{abstract}
  Single-frame atmospheric turbulence mitigation is inherently ill-posed due to spatially varying blur coupled with non-rigid geometric distortion. Existing end-to-end approaches trained on flat-field simulations often struggle to balance texture recovery with geometric rectification. To overcome this limitation, we propose D²Turb, a unified framework that bridges physics-grounded simulation with explicitly decoupled restoration. First, we introduce a Depth-Aware Turbulence Synthesis protocol that incorporates scene depth into the phase-to-space formulation. This generates physically consistent, depth-dependent degradations and provides a crucial intermediate tilt supervision signal for disentangled learning. Building upon this simulation engine, D²Turb decomposes restoration into two interactive stages: texture deblurring and geometric rectification. The texture deblurring stage employs a deblurring backbone to recover fine-grained details while preserving geometric distortion for the subsequent rectification stage. To mitigate the information fragmentation commonly observed in cascaded designs, we further propose an Adaptive Structural Prior Injection (ASPI) mechanism that dynamically transfers deep structural representations from the deblurring module to guide dense flow prediction for spatial unwarping. Extensive experiments demonstrate that D²Turb achieves state-of-the-art performance on both synthetic and real-world datasets, with consistent improvements in both texture recovery and geometric fidelity. Our code and pre-trained models are publicly available at \url{https://github.com/HertzDot222/D2Turb}.
\end{abstract}

\section{Introduction}
\label{sec:intro}

\begin{figure}[tb]
    \centering
    \includegraphics[width=\textwidth]{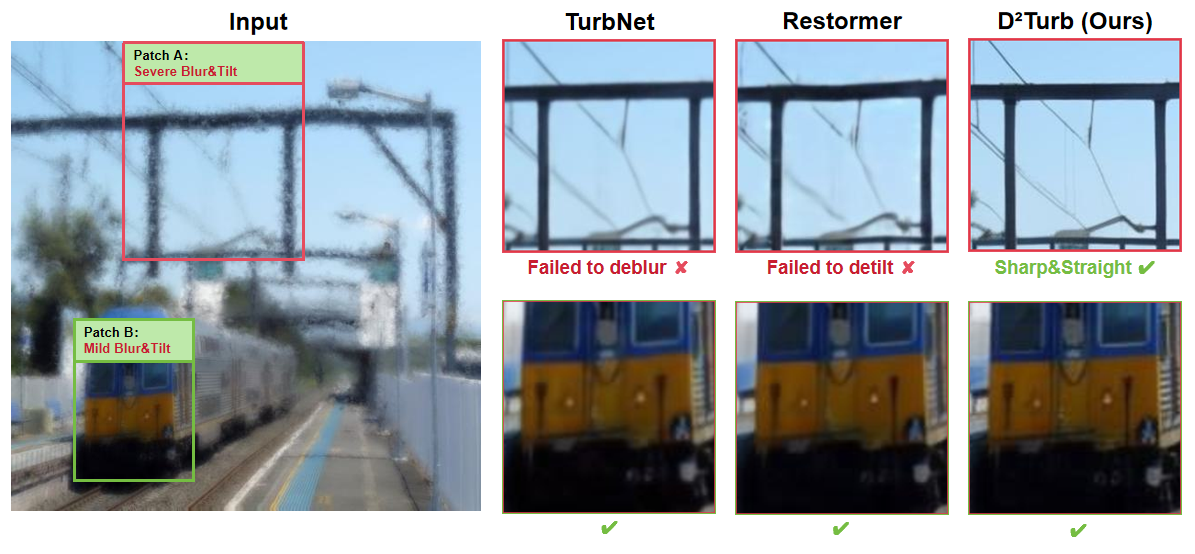} 
    \caption{\textbf{Motivation of D²Turb.} 
    \textbf{Left:} Real-world atmospheric turbulence exhibits inherent depth-dependent degradation. Here, using our physics-grounded simulator on a clean image (from Places365 \cite{zhou2017places}), we illustrate this phenomenon: distant regions suffer from severe blur and geometric warp (Patch A), while closer areas experience only mild effects (Patch B). 
    \textbf{Right:} While existing models handle mild degradation well, they suffer from a severe distortion-perception trade-off in highly degraded regions (Patch A). Specialized models (TurbNet) fail to deblur, and general backbones (Restormer) fail to unwarp. D²Turb uniquely recovers structures that are both sharp and straight.}
    \label{fig:motivation}
\end{figure}

Long-range imaging systems often suffer from severe degradation caused by atmospheric turbulence~\cite{roddier1981v,sutton2020atmospheric}. This phenomenon is driven by refractive-index fluctuations along the optical path~\cite{fried1966optical,tatarski1961wave}, manifesting as a complex entanglement of spatially varying blur and non-rigid geometric deformation (tilt) in the image domain. While multi-modal~\cite{xia2024nb, liu2025evturb,li2025egtm} or multi-frame methods~\cite{zhang2022tsr,zhang2024imaging,zhang2024spatio,xu2024long} exploit spatio-temporal statistics to mitigate these highly stochastic aberrations, their reliance on specialized hardware, strict temporal alignment, or static-scene assumptions drastically limits practical deployment. Consequently, pure-RGB single-frame restoration remains highly desirable but challenging.

Existing single-frame methods predominantly rely on monolithic, end-to-end architectures (\eg, AT-Net~\cite{yasarla2021learning}, TurbNet~\cite{mao2022single}, PiRN~\cite{jaiswal2023physics}) that attempt to jointly solve for blur and tilt. This entangled paradigm inevitably leads to a severe distortion-perception trade-off under complex degradations. Concurrently, general image restoration models (\eg, Restormer~\cite{zamir2022restormer}, FocalNet~\cite{cui2023focal}, AdaIR~\cite{cui2025adair}) have achieved remarkable success in recovering high-frequency details across various low-level vision tasks. While it is intuitive to directly deploy these powerful backbones for turbulence mitigation, they fundamentally lack explicit spatial mechanisms to unwarp severe non-rigid deformations. As clearly illustrated in Fig.~\ref{fig:motivation} (Patch A): specialized models struggle with deblurring, whereas general backbones recover sharpness but fail to unwarp. Furthermore, existing data-driven methods are severely bottlenecked by the physical fidelity of their training data. Current image-domain simulators, such as the Phase-to-Space (P2S) transform~\cite{mao2021accelerating}, apply uniform turbulence strength based on an idealized isoplanatic assumption. This ignores the path-accumulated, inherently depth-dependent nature of real-world turbulence, resulting in a profound sim-to-real domain gap.

To address these intertwined challenges, we propose D²Turb, a unified framework synergizing physics-grounded simulation with a decoupled restoration architecture. First, we introduce a Depth-Aware Turbulence Synthesis protocol that incorporates scene depth via the Kolmogorov power law~\cite{kolmogorov1991local}. This not only generates physically realistic data but uniquely provides an intermediate ``tilt'' ground truth. Leveraging this supervision, D²Turb explicitly disentangles restoration into texture deblurring and geometric rectification. To prevent information fragmentation between these cascaded stages, we devise an Adaptive Structural Prior Injection (ASPI) mechanism, which dynamically propagates deep structural semantics from the deblurring stage to guide dense flow prediction for spatial unwarping.

In summary, the main contributions of this work are four-fold:
\begin{itemize}
    \item \textbf{Physics-Grounded Synthesis:} We introduce a Depth-Aware Turbulence Synthesis protocol that physically models the path-accumulated nature of atmospheric degradation. This significantly narrows the sim-to-real domain gap and provides crucial ``tilt'' supervision for isolated geometric learning.
    \item \textbf{Explicitly Decoupled Architecture:} We propose D²Turb, a unified single-frame framework that fundamentally circumvents the distortion-perception trade-off by explicitly disentangling the restoration process into texture deblurring and geometric rectification.
    \item \textbf{Dynamic Feature Injection:} We devise the Adaptive Structural Prior Injection (ASPI) mechanism. By selectively fetching and spatially aligning deep structural semantics to guide dense flow prediction, ASPI fosters a mutual promotion between texture recovery and geometric unwarping.
    \item \textbf{New State-of-the-Art Baseline:} Extensive evaluations demonstrate our framework's superiority across synthetic and real-world datasets. Notably, D²Turb significantly outperforms existing methods, yielding a massive 19\% relative reduction in LPIPS and pushing the average PSNR to 25.72 dB on synthetic benchmarks, while setting new perceptual records on unannotated real-world captures. It establishes a robust, physically-grounded baseline to advance practical long-range imaging.
\end{itemize}

\section{Related Works}

\subsection{Single-Frame Turbulence Mitigation}
Single-frame atmospheric turbulence mitigation is fundamentally a highly ill-posed image restoration task. Modern general-purpose restoration models (\eg, FocalNet~\cite{cui2023focal}, Restormer~\cite{zamir2022restormer}, AdaIR~\cite{cui2025adair}) excel at reconstructing high-frequency textures. However, unlike conventional degradations, turbulence introduces spatially varying, non-rigid distortions tightly coupled with blur. Consequently, these appearance-driven models lack explicit spatial mechanisms to unwarp distorted regions, causing residual geometric artifacts. To address turbulence-specific characteristics, data-driven approaches like AT-Net~\cite{yasarla2021learning} employ cascaded sub-networks to estimate intermediate distortion masks, yet their end-to-end optimization remains fundamentally entangled. To introduce physical constraints, physics-inspired methods (\eg, TurbNet~\cite{mao2022single}) embed explicit degradation and reconstruction blocks. Despite their progress, these networks struggle to independently optimize texture and geometry priors, inevitably falling into a distortion-perception trade-off. Advanced transformer-based architectures like TMT~\cite{zhang2024imaging}, originally designed to explore disentanglement via temporal self-attention across multiple frames, can be adapted for single-frame training. Nevertheless, when deprived of temporal cues, they lack the explicit spatial mechanisms needed to decouple spatially varying blur from geometric warp.

\subsection{Atmospheric Turbulence Simulation}
Data-driven turbulence mitigation heavily relies on accurate simulation to generate paired training data. While 3D wave propagation models and split-step simulations~\cite{hardie2017simulation} are physically rigorous, they are computationally prohibitive for large-scale dataset generation. To address this, image-domain simulators, notably the Phase-to-Space (P2S) transform~\cite{chimitt2020simulating, mao2021accelerating}, efficiently map Zernike polynomial coefficients to spatially varying point spread functions (PSFs) and dense displacement fields. However, standard P2S simulators rely on an isoplanatic or flat-field assumption, typically applying a uniform turbulence strength across the entire image space. This deviates from real-world optical physics, where turbulence intensity accumulates dynamically along the line of sight, resulting in depth-independent synthesized degradations and a severe sim-to-real domain gap for models trained on such data.

\section{Methodology}

\begin{figure}[tb]
    \centering
    \includegraphics[width=\textwidth]{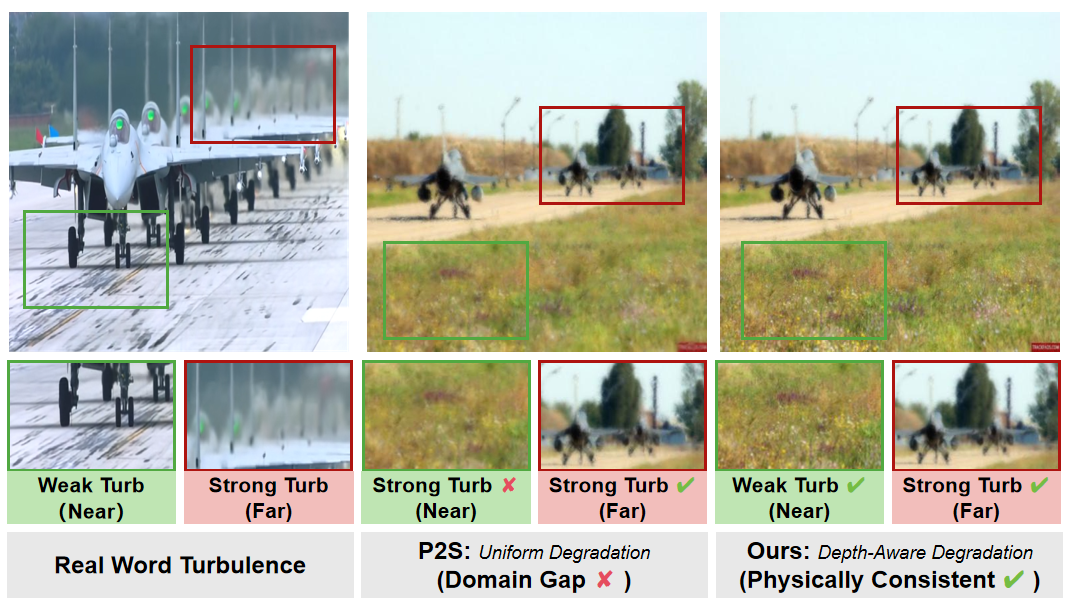} 
    \caption{\textbf{Sim-to-real domain gap in turbulence simulation.} 
    \textbf{Left:} Real-world turbulence accumulates over the optical path, preserving foregrounds (green) while severely distorting distant objects (red). 
    \textbf{Middle:} Standard P2S incorrectly applies uniform degradation across all depths. 
    \textbf{Right:} Our Depth-Aware Synthesis accurately replicates physically consistent, depth-dependent degradations.}
    \label{fig:sim-P2SVSOurs}
\end{figure}

\begin{figure}[tb]
    \centering
    \includegraphics[width=\textwidth]{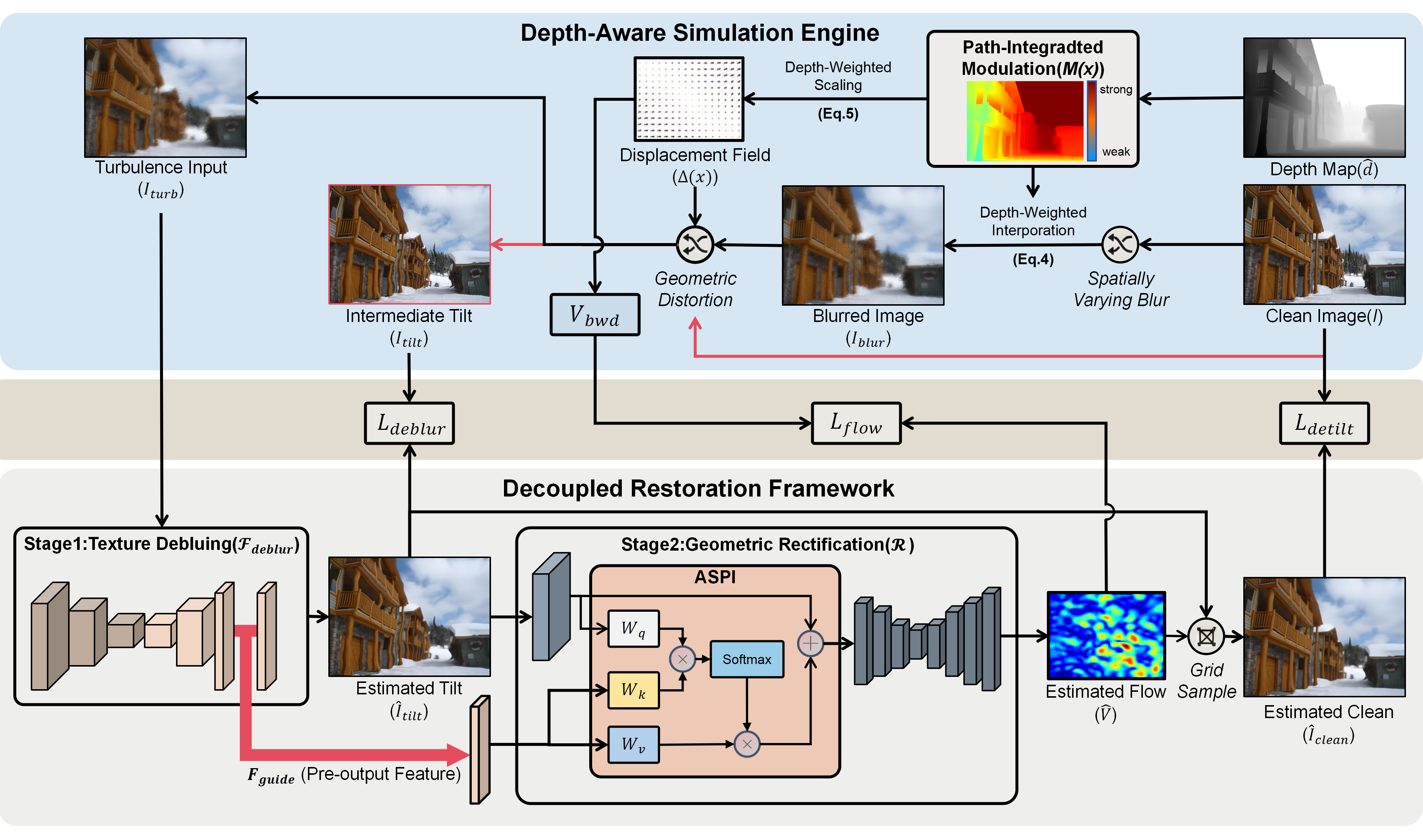} 
    \caption{\textbf{Overview of the proposed D²Turb framework.} 
    \textbf{Top:} The Depth-Aware Simulation Engine synthesizes physically consistent turbulence ($I_{turb}$) and an intermediate ``tilt'' ground truth ($I_{tilt}$, red path) by explicitly modulating blur and distortion via a path-integrated depth map ($M(x)$). 
    \textbf{Middle:} Multi-stage supervision effectively closes the training loop. 
    \textbf{Bottom:} In our decoupled restoration framework, Stage 1 ($\mathcal{F}_{deblur}$) recovers texture and extracts deep structural priors ($F_{guide}$). Stage 2 ($\mathcal{R}$) utilizes Adaptive Structural Prior Injection (ASPI) to dynamically fuse $F_{guide}$ to predict a dense flow field ($\hat{\mathcal{V}}$) for precise spatial unwarping.}
    \label{fig:framework}
\end{figure}

\subsection{Physics-Grounded Simulation Engine}
To narrow the sim-to-real domain gap (Fig.~\ref{fig:sim-P2SVSOurs}), we propose a Depth-Aware Turbulence Synthesis protocol grounded in physical optics (Fig.~\ref{fig:framework}, top). Let $I \in \mathbb{R}^{H \times W \times 3}$ denote a clean image and $\tilde{d} \in [0, 1]^{H \times W}$ be its relative depth map. We first project $\tilde{d}$ to a physical propagation distance $z(x)$ at spatial location $x$:
\begin{equation}
z(x) = L \cdot \left((1 - s) \cdot \tilde{d}(x) + s\right),
\end{equation}
where $L$ is the total optical path length, and $s \in (0, 1)$ acts as a scene-specific baseline offset. Since $\tilde{d} \in [0, 1]$, directly utilizing it would erroneously map the closest visible object to zero distance, generating no turbulence. However, in long-range telephoto imaging, even foreground objects remain distant from the camera and are still affected by turbulence. By introducing $s$, we flexibly calibrate the dynamic range of the physical distance (\eg, mapping a distant, narrow-depth scene to $[0.9L, L]$), ensuring the simulated degradation accurately reflects the authentic path-accumulated turbulence strength.

According to Kolmogorov turbulence theory~\cite{kolmogorov1991local}, assuming a uniform refractive index structure constant $C_n^2$ along the horizontal path, the Fried parameter scales with distance following a $\frac{3}{5}$ power law:
\begin{equation}
    r_0(z) = r_0(L) \cdot \left(\frac{L}{z}\right)^{\frac{3}{5}}.
\end{equation}
Building upon this, we formulate a physically rigorous, spatially varying depth modulation map $M(x) \in [0, 1]$, normalized by the maximum accumulated turbulence strength $z_{max}$:
\begin{equation}
    M(x) = \left(\frac{z(x)}{z_{max}}\right)^{\frac{3}{5}}.
\end{equation}
This explicitly replaces the non-physical uniform mapping ($M(x) \equiv 1$) of standard simulators. Guided by $M(x)$, the depth-aware blur is formulated as a spatial interpolation:
\begin{equation}
    I_{blur}(x) = M(x) \cdot (\mathcal{K}(x) * I(x)) + (1 - M(x)) \cdot I(x),
\end{equation}
where $\mathcal{K}(x)$ represents the spatially varying PSF generated from the Zernike polynomial basis. Similarly, the depth-aware geometric distortion is modulated by scaling the phase-derived pixel-wise displacement vector:
\begin{equation}
    \Delta(x) = M(x) \cdot \mathcal{F}^{-1}(\Phi),
\end{equation}
where $\Phi$ denotes the frequency-domain phase aberration screen governed by Kolmogorov turbulence statistics, and $\mathcal{F}^{-1}(\cdot)$ represents the inverse Fourier transform operation utilized to map the phase representations into the spatial dense displacement field~\cite{mao2021accelerating}.
The final turbulent image is generated by applying this dense field to the blurred image:
\begin{equation}
    I_{turb}(x) = I_{blur}(x + \Delta(x)).
\end{equation}

Crucially, to enable explicitly decoupled training, our simulator circumvents the blur module ($I_{blur} \equiv I$) to produce an intermediate ``tilt'' ground truth:
\begin{equation}
I_{tilt}(x) = I(x + \Delta(x)).
\end{equation}
This isolates geometric distortion from texture blur, providing precise structural supervision for the subsequent disentangled learning stages.

\subsection{The D²Turb Framework}
Leveraging this unique physical supervision, D²Turb explicitly disentangles restoration into two interactive stages (Fig.~\ref{fig:framework}, bottom).

\subsubsection{Texture Prior Extraction via Decoupled Deblurring:}
To recover high-frequency textures from $I_{turb}$ and predict the intermediate state $I_{tilt}$, we employ a deblurring module ($\mathcal{F}_{deblur}$) optimized jointly from scratch:
\begin{equation}
    \hat{I}_{tilt}, F_{guide} = \mathcal{F}_{deblur}(I_{turb}).
\end{equation}
Directed strictly by the ``tilt'' supervision, $\mathcal{F}_{deblur}$ adapts to spatially varying blur without prematurely correcting geometric deformations. Since most modern networks predict RGB outputs via simple linear projections from deep latent spaces, we can flexibly instantiate $\mathcal{F}_{deblur}$ using any off-the-shelf backbone (\eg, Restormer~\cite{zamir2022restormer}, FocalNet~\cite{cui2023focal}) by extracting its pre-output feature $F_{guide} \in \mathbb{R}^{H \times W \times C}$. This architecture-agnostic design circumvents the severe information bottleneck of standard cascades, which pass only the compressed 3-channel $\hat{I}_{tilt}$. Consequently, $F_{guide}$ preserves a rich, high-dimensional embedding of uncorrupted structural priors, providing an expressive key-value space for the subsequent ASPI mechanism to guide precise geometric rectification.

\subsubsection{Geometric Rectification via Dense Flow Prediction:}
To address non-rigid distortions, we construct a Tilt Rectifier ($\mathcal{R}$), implemented as a lightweight U-Net~\cite{ronneberger2015u} (architectural details are provided in the Supplementary Material), to predict a dense, pixel-level displacement field. Given an enriched input feature, $\mathcal{R}$ predicts a backward flow field $\hat{\mathcal{V}} \in \mathbb{R}^{H \times W \times 2}$. A parameter-free, differentiable grid sampling operation ($\mathcal{S}$) then warps the intermediate image to generate the final restored output:
\begin{equation}
    \hat{I}_{clean} = \mathcal{S}(\hat{I}_{tilt}, \hat{\mathcal{V}}).
\end{equation}

\subsubsection{Adaptive Structural Prior Injection (ASPI):}
A naive cascaded approach, which directly feeds the compressed $\hat{I}_{tilt}$ into $\mathcal{R}$, suffers from severe spatial misalignment and information fragmentation. To bridge this semantic gap, our ASPI mechanism leverages the expressive key-value space of $F_{guide}$ to dynamically guide the shallow geometric embedding $F_{shallow} = \mathcal{E}(\hat{I}_{tilt})$ of the rectifier. Formulated as a cross-attention process, the fused feature is computed by:
\begin{equation}
    F_{fused} = \text{Softmax}\left(\frac{Q K^T}{\sqrt{d}}\right)V + F_{shallow},
\end{equation}
where queries derive from the geometric embedding ($Q = W_q F_{shallow}$), keys and values project from the deep texture prior ($K = W_k F_{guide}$, $V = W_v F_{guide}$), and $d$ is the channel dimension. This attention mechanism empowers the network to selectively fetch sharp edge topologies from $F_{guide}$ specifically at severely warped regions. Furthermore, gradients from the final rectification loss flow back through $F_{guide}$, explicitly forcing the upstream deblurring module to preserve geometries conducive to spatial unwarping, thereby establishing a closed-loop mutual promotion.

\subsection{Loss Functions}
D²Turb is optimized end-to-end using a tightly coupled multi-stage objective. To enforce architectural decoupling, the first stage is strictly supervised by the ``tilt'' ground truth:
\begin{equation}
    \mathcal{L}_{deblur} = \mathcal{L}_{\mathcal{F}_{deblur}}(\hat{I}_{tilt}, I_{tilt}),
\end{equation}
where $\mathcal{L}_{\mathcal{F}_{deblur}}$ denotes the native reconstruction objective of the chosen deblurring module. 

For geometric restoration, since our simulator mathematically generates a forward grid, we employ a differentiable forward splatting algorithm (details are provided in the Supplementary Material) to invert it into an accurate backward flow field $\mathcal{V}_{bwd}$. The dense displacement prediction is then regularized by:
\begin{equation}
    \mathcal{L}_{flow} = \|\hat{\mathcal{V}} - \mathcal{V}_{bwd}\|_1.
\end{equation}

Finally, to capture both photometric correctness and structural perception against the clean ground truth $I_{gt}$, the rectification constraint $\mathcal{L}_{detilt}$ is formulated as:
\begin{equation}
    \mathcal{L}_{detilt} = \lambda_{photo}\|\hat{I}_{clean} - I_{gt}\|_1 + \lambda_{vgg}\|\phi(\hat{I}_{clean}) - \phi(I_{gt})\|_2^2,
\end{equation}
where $\phi(\cdot)$ extracts pre-trained VGG~\cite{simonyan2014very} feature maps. The total framework loss is optimized as:
\begin{equation}
    \mathcal{L}_{total} = \mathcal{L}_{deblur} + \lambda_{flow}\mathcal{L}_{flow} + \mathcal{L}_{detilt}.
\end{equation}

Here, $\lambda_{photo}$, $\lambda_{vgg}$, and $\lambda_{flow}$ are hyperparameter weights designed to balance the gradient magnitudes of pixel-level photometric fidelity, high-frequency perceptual similarity, and kinematic constraints, respectively. To prevent any single objective from dominating the joint optimization, they are set to $\lambda_{photo} = 1.0$, $\lambda_{vgg} = 0.02$, and $\lambda_{flow} = 5.0$ in our implementation.

\section{Experiments}

\subsection{Experimental Setup}
\subsubsection{Datasets and Implementation Details:}
Using Places365~\cite{zhou2017places} and Depth Anything V2~\cite{yang2024depth}, our engine synthesizes 88,000 training and 15,200 testing pairs. The synthetic testing set is categorized by turbulence strength $D/r_0$ into weak ($<2.25$), medium ($[2.25, 3.75]$), and strong ($>3.75$) subsets. We also evaluate on the real-world RLR-AT dataset~\cite{xu2024long} to validate sim-to-real generalization. Implemented in PyTorch, D²Turb is trained from scratch for 50 epochs using the Adam optimizer (cosine annealing learning rate from $1 \times 10^{-4}$ to $1 \times 10^{-6}$). Models are trained on four RTX 4090 GPUs using $256 \times 256$ patches, a total batch size of 8. All baselines are retrained identically for strictly fair comparison. Detailed simulation configurations are provided in the Supplementary Material.

\subsubsection{Evaluation Metrics:}
For the synthetic test set, we report PSNR, SSIM~\cite{wang2004image}, and LPIPS~\cite{zhang2018unreasonable}. For the unannotated RLR-AT dataset, we utilize robust no-reference metrics: NIQE~\cite{mittal2012making} and MUSIQ~\cite{ke2021musiq}.

\subsection{Depth-Aware Simulation}

\begin{table}[tb]
    \caption{Quantitative sim-to-real evaluation on the real-world RLR-AT dataset. TurbNet trained on our depth-aware dataset outperforms the P2S baseline.}
    \label{tab:sim2real}
    \centering
    \begin{tabularx}{\linewidth}{@{} l *{2}{>{\centering\arraybackslash}X} @{}}
    \toprule
         Training Data (Simulator) & NIQE$\downarrow$ & MUSIQ$\uparrow$ \\
    \midrule
         Standard P2S & 7.397 & 46.554 \\
         \textbf{Ours (Depth-Aware)} & \textbf{6.980} & \textbf{51.996} \\
    \bottomrule
    \end{tabularx}
\end{table}

\begin{figure}[tb]
    \centering
    \includegraphics[width=1.0\linewidth]{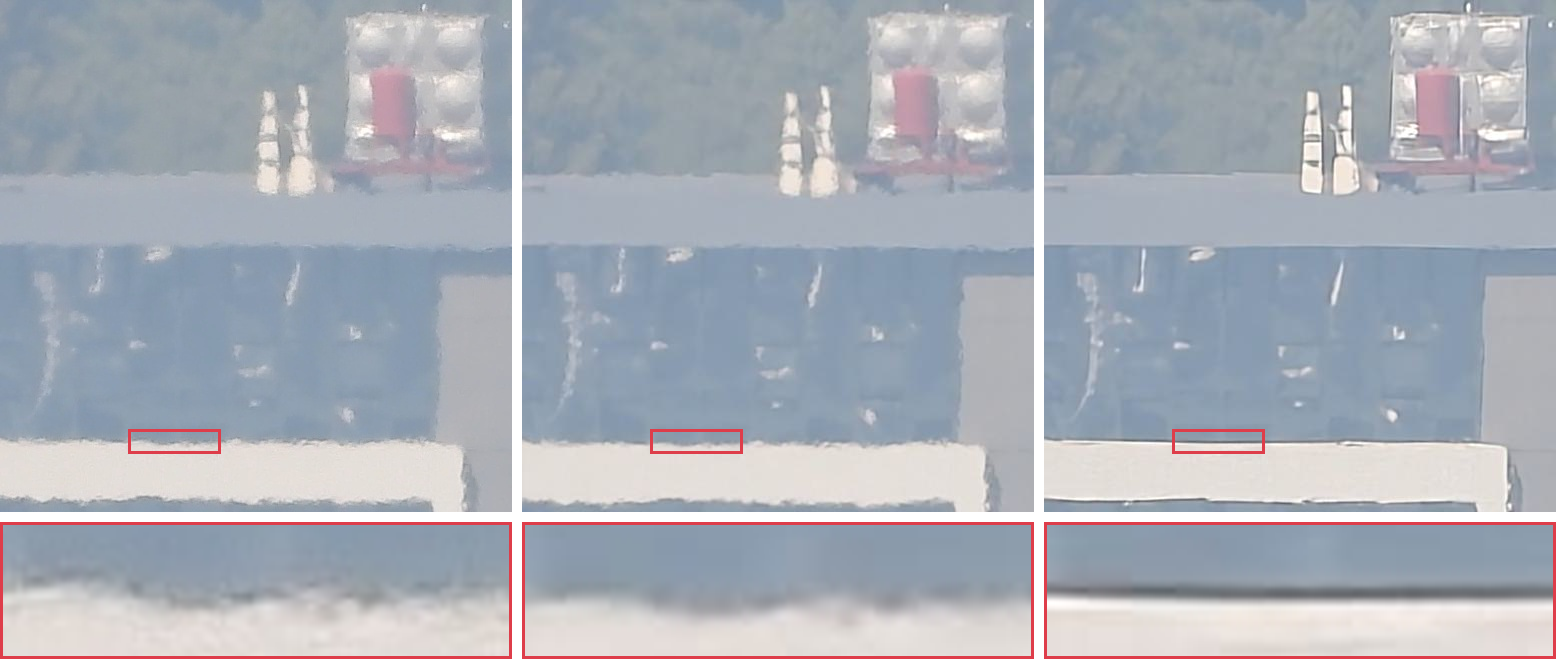}
    \scriptsize
    \makebox[0.333\linewidth]{Input}%
    \makebox[0.333\linewidth]{Trained With P2S}%
    \makebox[0.333\linewidth]{\textbf{Trained With Ours}}%
    \caption{\textbf{Sim-to-real visual comparison.} The P2S-trained model struggles with complex depth-varying degradations. Our depth-aware training effectively straightens warped edges and recovers authentic textures.}
    \label{fig:sim2real_visual}
\end{figure}

To validate our depth-aware synthesis in bridging the sim-to-real gap, we compare it against the uniform P2S simulator using two parallel TMT datasets~\cite{zhang2024imaging} with identical physical parameters. When training the TurbNet baseline from scratch on both, our depth-modulated data significantly outperforms the P2S counterpart on the real-world RLR-AT dataset~\cite{xu2024long} across no-reference metrics (Table \ref{tab:sim2real}). Visually (Fig.~\ref{fig:sim2real_visual}), our synthesis provides superior structural supervision, effectively suppressing the residual geometric distortions left by flat-field training. Detailed dataset configurations and ablations regarding the physical $3/5$ power law are provided in the Supplementary Material.

\subsection{Comparison with State-of-the-Art Methods}
We compare D²Turb (equipped with the Restormer backbone) against specialized single-frame methods (AT-Net~\cite{yasarla2021learning}, TurbNet~\cite{mao2022single}, TMT(single-frame Input)~\cite{zhang2024imaging}) and advanced general restoration backbones (FocalNet~\cite{cui2023focal}, Restormer~\cite{zamir2022restormer}, AdaIR~\cite{cui2025adair}). 

\begin{table}[tb]
\caption{Quantitative comparison on the synthetic test set across weak ($D/r_0 < 2.25$), medium ($2.25 \le D/r_0 \le 3.75$), and strong ($D/r_0 > 3.75$) turbulence intensities. The best and second-best results are highlighted in \textbf{bold} and \underline{underline}, respectively.}
\label{tab:sota_synthetic}
\centering
\resizebox{\linewidth}{!}{
\begin{tabular}{@{} l *{12}{c} @{}}
\toprule
\multirow{2}{*}{Method} & \multicolumn{4}{c}{PSNR$\uparrow$} & \multicolumn{4}{c}{SSIM$\uparrow$} & \multicolumn{4}{c}{LPIPS$\downarrow$} \\
\cmidrule(lr){2-5} \cmidrule(lr){6-9} \cmidrule(l){10-13}
& weak & medium & strong & \textbf{Avg} & weak & medium & strong & \textbf{Avg} & weak & medium & strong & \textbf{Avg} \\
\midrule
AT-Net & 26.869 & 24.529 & 23.287 & 24.904 & 0.795 & 0.697 & 0.640 & 0.711 & 0.196 & 0.299 & 0.378 & 0.291 \\
TurbNet & \underline{27.513} & \underline{25.174} & \underline{24.010} & \underline{25.575} & 0.805 & 0.713 & 0.658 & 0.726 & 0.198 & 0.298 & 0.375 & 0.290 \\
TMT & 27.413 & 24.992 & 23.767 & 25.400 & 0.805 & 0.712 & 0.658 & 0.726 & 0.189 & 0.270 & 0.334 & 0.262 \\
\midrule
FocalNet & 27.459 & 25.101 & 23.922 & 25.503 & 0.805 & 0.711 & 0.656 & 0.724 & 0.198 & 0.302 & 0.380 & 0.292 \\
Restormer & 27.499 & 25.112 & 23.922 & 25.521 & 0.808 & 0.717 & 0.664 & 0.730 & \underline{0.179} & \underline{0.265} & \underline{0.330} & \underline{0.257} \\
AdaIR & 27.510 & 25.125 & 23.936 & 25.533 & \underline{0.808} & \underline{0.717} & \underline{0.664} & \underline{0.730} & 0.179 & 0.266 & 0.331 & 0.258 \\
\midrule
\textbf{D²Turb (Ours)} & \textbf{27.660} & \textbf{25.331} & \textbf{24.153} & \textbf{25.724} & \textbf{0.811} & \textbf{0.724} & \textbf{0.673} & \textbf{0.736} & \textbf{0.150} & \textbf{0.215} & \textbf{0.261} & \textbf{0.208} \\
\bottomrule
\end{tabular}
}
\end{table}

\begin{figure}[tb]
    \centering
    \includegraphics[width=1.0\linewidth]{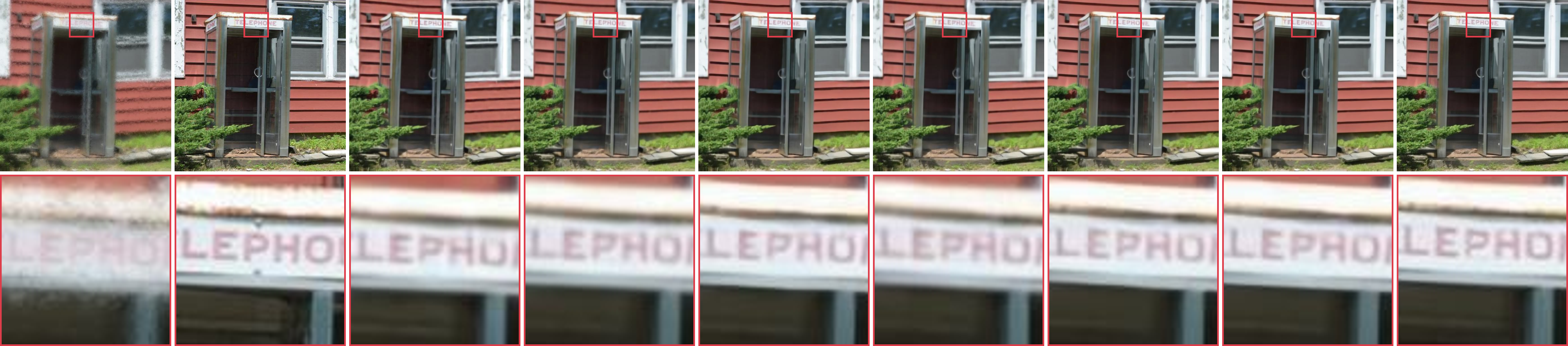} \\
    \vspace{1mm} 
    \includegraphics[width=1.0\linewidth]{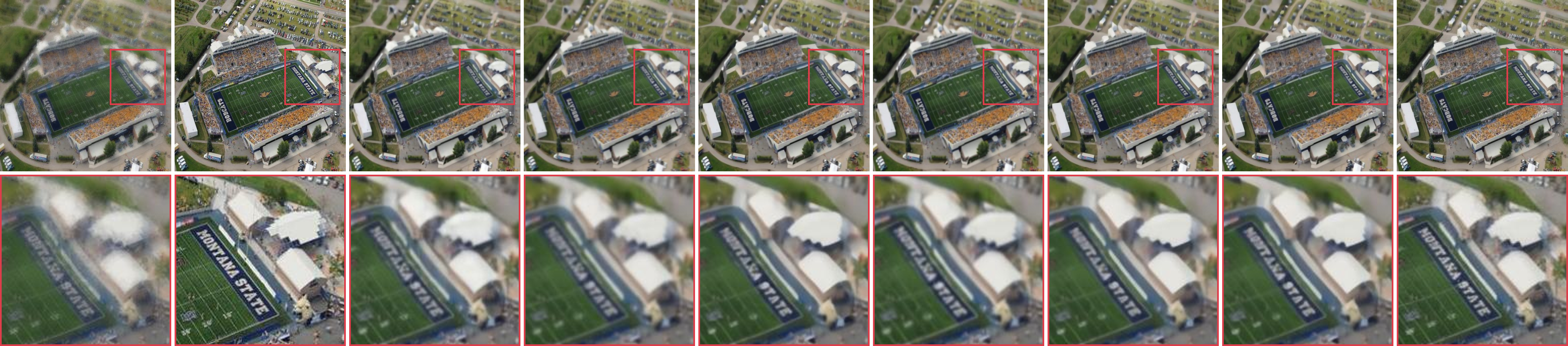} \\
    \vspace{1mm} 
    \includegraphics[width=1.0\linewidth]{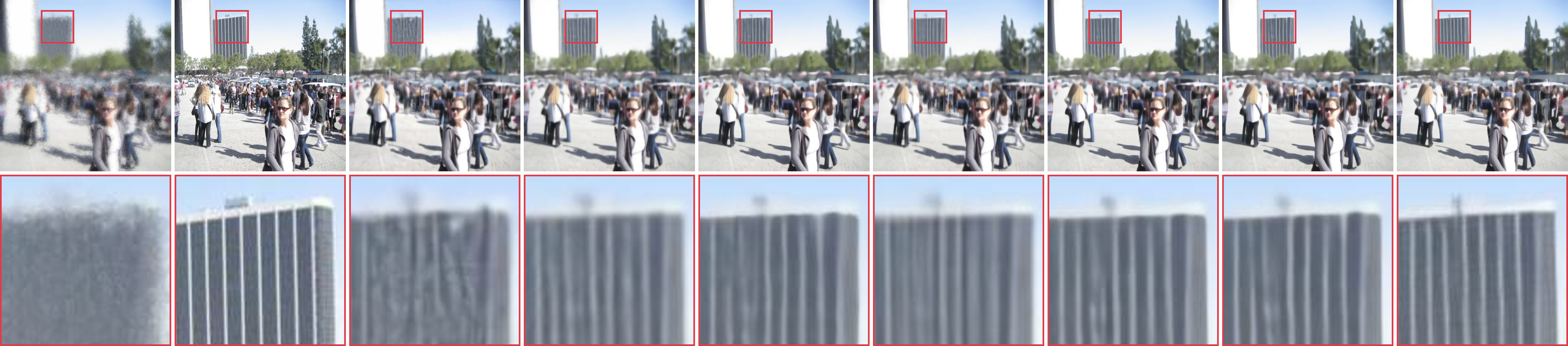} \\
    \vspace{1mm} 
    \scriptsize
    \makebox[0.111\linewidth]{Input}%
    \makebox[0.111\linewidth]{GT}%
    \makebox[0.111\linewidth]{AT-Net}%
    \makebox[0.111\linewidth]{TurbNet}%
    \makebox[0.111\linewidth]{TMT}%
    \makebox[0.111\linewidth]{FocalNet}%
    \makebox[0.111\linewidth]{Restormer}%
    \makebox[0.111\linewidth]{AdaIR}%
    \makebox[0.111\linewidth]{\textbf{Ours}}
    
    \caption{\textbf{Qualitative comparison on synthetic data.} General restoration models (\eg, Restormer, AdaIR) recover sharp textures but fail to correct non-rigid warping (evidenced by the wavy ``TELEPHONE'' text and building facades). Specialized turbulence models (\eg, TurbNet, TMT) attempt geometric correction but severely over-smooth high-frequency details. By explicitly decoupling these tasks, D²Turb uniquely recovers highly legible text and perfectly straight geometric contours without hallucination. It is worth noting that a certain degree of spatial misalignment between the degraded inputs and the clean ground truth is inherently expected, as it reflects the severe, non-rigid geometric distortions naturally induced by atmospheric turbulence.}
    \label{fig:syn_visual}
\end{figure}

\begin{table}[tb]
    \caption{Quantitative sim-to-real evaluation on the real-world RLR-AT dataset. Performance is measured using no-reference perceptual metrics. The best and second-best results are highlighted in \textbf{bold} and \underline{underline}, respectively.}
    \label{tab:real_world}
    \centering
    \begin{tabularx}{\linewidth}{@{} l *{2}{>{\centering\arraybackslash}X} @{}}
    \toprule
         Method & NIQE$\downarrow$ & MUSIQ$\uparrow$ \\
    \midrule
         AT-Net  & \underline{6.908} & 48.365 \\
         TurbNet  & 7.169 & 48.387 \\
         TMT  & 7.049 & \underline{52.418} \\
    \midrule
         FocalNet  & 7.285 & 46.471 \\
         Restormer  & 6.995 & 50.705 \\
         AdaIR  & 6.933 & 51.135 \\
    \midrule
         \textbf{D²Turb (Ours)} & \textbf{6.653} & \textbf{52.815} \\
    \bottomrule
    \end{tabularx}
\end{table}

\begin{figure}[tb]
    \centering
    \includegraphics[width=1.0\linewidth]{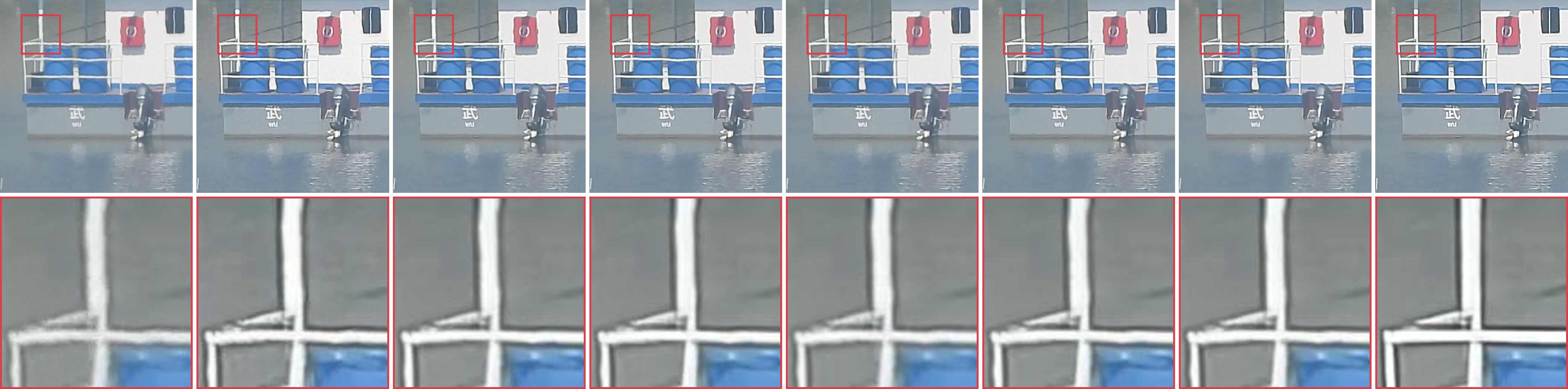} \\
    \vspace{1mm}
    \includegraphics[width=1.0\linewidth]{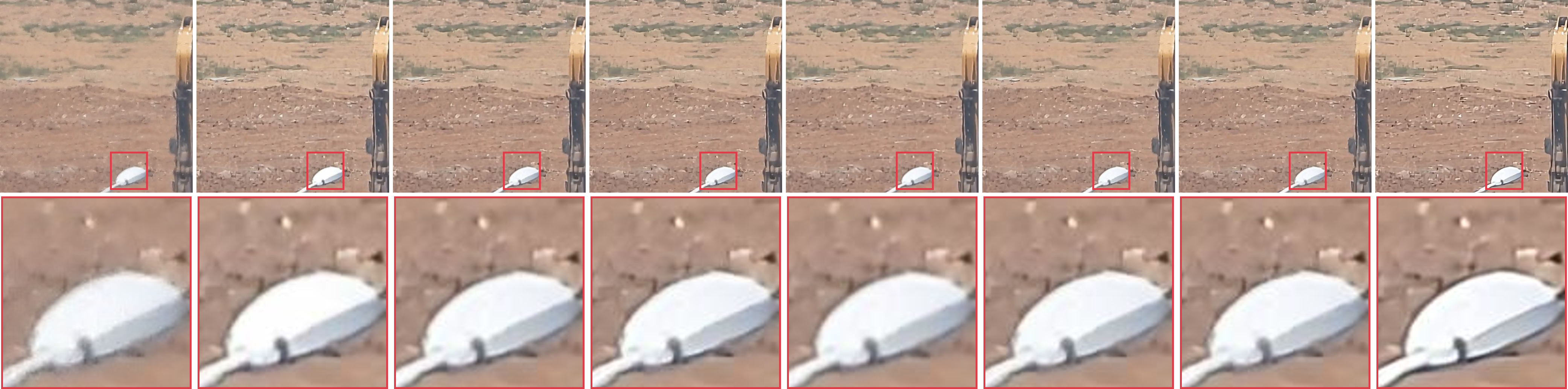} \\
    \vspace{1mm}
    \includegraphics[width=1.0\linewidth]{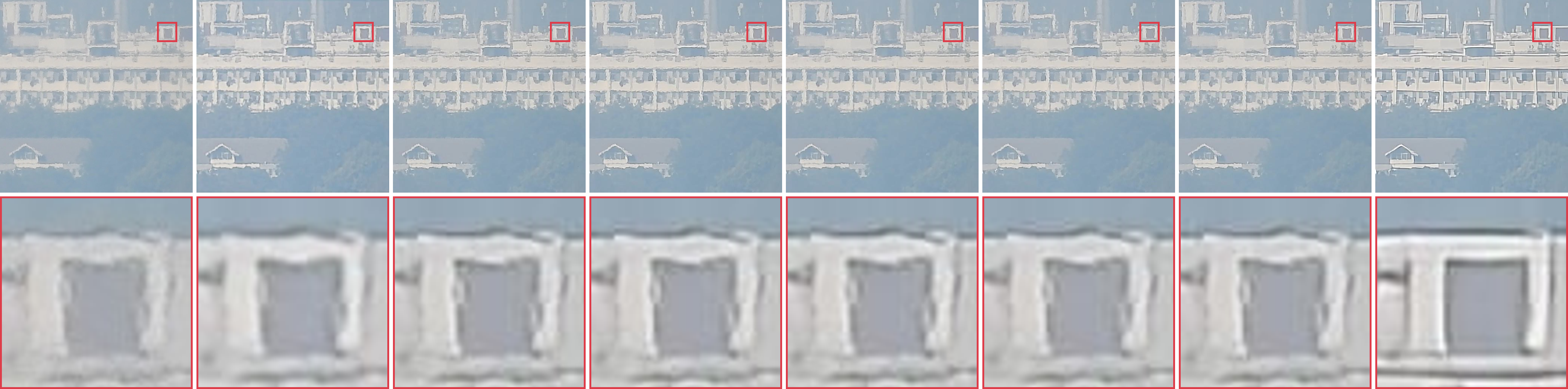} \\
    \vspace{1mm}
    \scriptsize
    \makebox[0.125\linewidth]{Input}%
    \makebox[0.125\linewidth]{AT-Net}%
    \makebox[0.125\linewidth]{TurbNet}%
    \makebox[0.125\linewidth]{TMT}%
    \makebox[0.125\linewidth]{FocalNet}%
    \makebox[0.125\linewidth]{Restormer}%
    \makebox[0.125\linewidth]{AdaIR}%
    \makebox[0.125\linewidth]{\textbf{Ours}}
    \caption{\textbf{Qualitative comparison on real-world data.} Competing methods fail to register non-rigid displacements or produce unnatural structural tears. D²Turb consistently unwarps severe, path-accumulated deformations.}
    \label{fig:real_visual}
\end{figure}

\subsubsection{Evaluation on Synthetic Data:}
As shown in Table \ref{tab:sota_synthetic}, D²Turb consistently achieves the best overall performance, pushing the average PSNR to 25.724 dB. Notably, it yields a massive 19\% relative reduction in LPIPS (0.208). Since LPIPS heavily penalizes both structural misalignments (tilt) and detail loss (blur), this perceptual breakthrough directly validates our explicit decoupling paradigm. As visually corroborated in Fig.~\ref{fig:syn_visual}, while general backbones fail to unwarp and specialized models over-smooth details, D²Turb uniquely restores text legibility and rigid contours without hallucination.

\subsubsection{Evaluation on Real-World Data:}
We rigorously assess sim-to-real generalization on the authentic RLR-AT dataset. As shown in Table \ref{tab:real_world}, D²Turb significantly outperforms all competing methods across perceptual scores. This is visually supported in Fig.~\ref{fig:real_visual}. General restoration models remain highly vulnerable to unconstrained turbulence without explicit spatial unwarping. Competing specialized methods either fail to register complex displacements or produce ghosting artifacts. Empowered by depth-aware training and explicitly decoupled learning, D²Turb consistently unwarps severe deformations.

\subsection{Ablation Study}

We conduct extensive ablation studies using our optimal configuration (D²Turb with the Restormer backbone). Consistent performance trends across other state-of-the-art backbones (\eg, FocalNet and AdaIR) are provided in the Supplementary Material.

\subsubsection{Effectiveness of the ASPI Mechanism:}

\begin{table}[tb]
  \caption{Ablation study on the feature injection mechanism. Our proposed ASPI effectively resolves spatial misalignment, yielding the best structural fidelity on the synthetic test set and the highest perceptual quality on the real-world dataset.}
  \label{tab:ablation_aspi}
  \centering
  \begin{tabularx}{\linewidth}{@{} l *{5}{>{\centering\arraybackslash}X} @{}}
    \toprule
    \multirow{2}{*}{\textbf{Variant (Injection Type)}} & \multicolumn{3}{c}{\textbf{Synthetic Dataset}} & \multicolumn{2}{c}{\textbf{Real-World Dataset}} \\
    \cmidrule(lr){2-4} \cmidrule(lr){5-6}
    & PSNR$\uparrow$ & SSIM$\uparrow$ & LPIPS$\downarrow$ & NIQE$\downarrow$ & MUSIQ$\uparrow$ \\
    \midrule
    Pure Cascade (w/o Injection) & 25.417 & 0.725 & 0.225 & 6.845 & 51.842 \\
    Simple Concat                & 25.504 & 0.730 & 0.221 & 6.921 & 52.520 \\
    ASPI (Ours)                  & \textbf{25.724} & \textbf{0.736} & \textbf{0.208} & \textbf{6.653} & \textbf{52.815} \\
    \bottomrule
  \end{tabularx}
\end{table}

\begin{figure}[tb]
    \centering
    \includegraphics[width=1.0\linewidth]{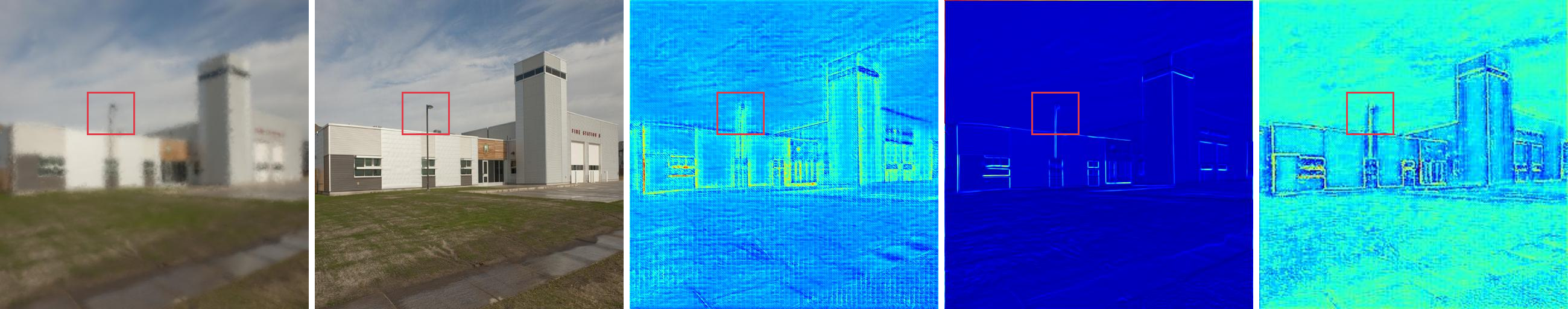} \\
    \vspace{1mm} 
    \includegraphics[width=1.0\linewidth]{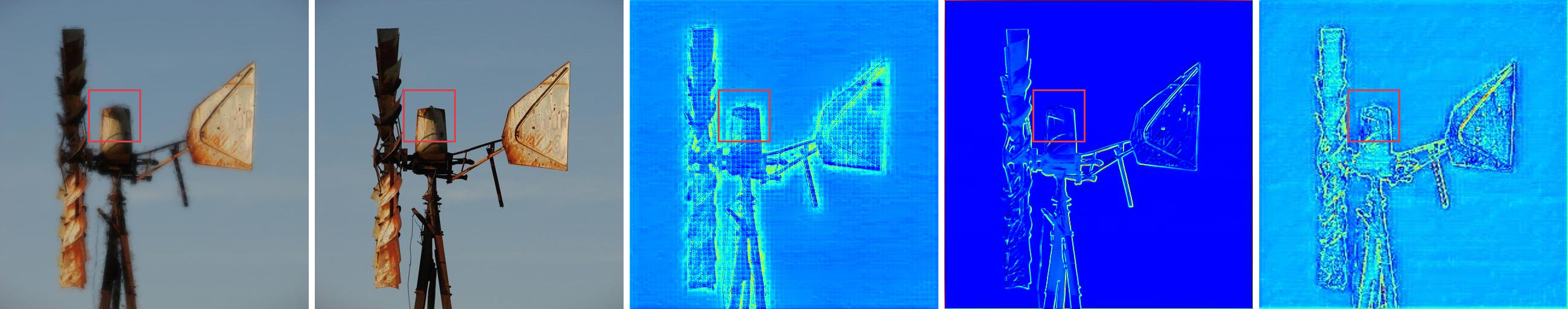} \\
    \vspace{1mm} 
    \scriptsize
    \makebox[0.2\linewidth]{Input}%
    \makebox[0.2\linewidth]{Ground Truth}%
    \makebox[0.2\linewidth]{$F_{guide}$}%
    \makebox[0.2\linewidth]{w/o ASPI}%
    \makebox[0.2\linewidth]{w/ ASPI}
    \caption{\textbf{Feature visualization of the ASPI mechanism.} We visualize intermediate representations extracted from specific topological stages of the framework. The raw prior ($F_{guide}$) is explicitly extracted as the pre-output feature from the upstream texture deblurring module. The unmodulated map (w/o ASPI) represents the intermediate feature state immediately before entering the ASPI module. Conversely, the modulated map (w/ ASPI) visualizes the direct output feature of the ASPI module, where spatially aligned edge priors (highlighted by red boxes) are strongly activated to guide the subsequent geometric rectification.}
    \label{fig:aspi_feat}
\end{figure}

The critical challenge in a decoupled pipeline lies in effectively transferring high-frequency structural priors to the geometric rectifier. We investigate three integration variants: (1) Pure Cascade (passing only the intermediate state without deep priors), (2) Simple Concat (naive channel-wise concatenation), and (3) our proposed ASPI.

As reported in Table \ref{tab:ablation_aspi}, the pure cascade approach suffers from severe information fragmentation. While simple concatenation yields marginal gains, it blindly passes uniform structural priors across the spatial domain. This lack of spatial focus overwhelms the downstream rectifier and exacerbates local misalignments, resulting in sub-optimal real-world perceptual scores.

Fig.~\ref{fig:aspi_feat} visually demystifies our solution. The raw structural prior ($F_{guide}$) contains rich high-frequency textures but lacks specific spatial focus. Without our injection mechanism (w/o ASPI), the geometric rectifier exhibits extremely weak and sparse activations, failing to locate structural boundaries. In contrast, ASPI operates as a dynamic spatial attention mechanism, empowering the network to selectively fetch and align sharp edge topologies from $F_{guide}$. As highlighted by the red boxes (\eg, the thin pole), the feature map after injection (w/ ASPI) displays highly concentrated activations specifically at severely warped regions. By dynamically gating these targeted priors, ASPI resolves spatial misalignments and ensures the highest structural fidelity.

\subsubsection{Effectiveness of Multi-Stage Supervision:}

\begin{table}[tb]
\caption{Ablation study on the multi-stage loss functions (using the Restormer backbone). Both the intermediate structural supervision ($\mathcal{L}_{deblur}$) and the geometric kinematic constraint ($\mathcal{L}_{flow}$) are indispensable for the explicitly decoupled framework, consistently improving both synthetic fidelity and real-world perceptual quality.}
\label{tab:ablation_loss}
\centering
\begin{tabularx}{\linewidth}{@{} l *{5}{>{\centering\arraybackslash}X} @{}}
\toprule
\multirow{2}{*}{\textbf{Loss Configuration}} & \multicolumn{3}{c}{\textbf{Synthetic Dataset}} & \multicolumn{2}{c}{\textbf{Real-World Dataset}} \\
\cmidrule(lr){2-4} \cmidrule(l){5-6}
& PSNR$\uparrow$ & SSIM$\uparrow$ & LPIPS$\downarrow$ & NIQE$\downarrow$ & MUSIQ$\uparrow$ \\
\midrule
w/o $\mathcal{L}_{\text{deblur}}$ & 25.536 & 0.725 & 0.221 & 6.866 & 52.455 \\
w/o $\mathcal{L}_{\text{flow}}$   & 25.372 & 0.722 & 0.229 & 6.827 & 51.498 \\
\textbf{Full Objective}    & \textbf{25.724} & \textbf{0.736} & \textbf{0.208} & \textbf{6.653} & \textbf{52.815} \\
\bottomrule
\end{tabularx}
\end{table}

To validate the necessity of our specifically designed loss functions, we ablate the intermediate deblurring loss ($\mathcal{L}_{deblur}$) and the inverse kinematics flow loss ($\mathcal{L}_{flow}$). As reported in Table \ref{tab:ablation_loss}, removing $\mathcal{L}_{deblur}$ causes the framework to lose its explicit intermediate "tilt" supervision, effectively degrading the disentangled architecture into an entangled end-to-end mapping and severely hurting structural fidelity. Similarly, discarding $\mathcal{L}_{flow}$ deprives the geometric rectifier of vital kinematic constraints, preventing accurate spatial unwarping. The full multi-stage objective yields the best performance, ensuring rigorous decoupling and geometric accuracy.

\subsubsection{Analysis of Mutual Promotion:}

\begin{table}[tb]
\caption{Analysis of Mutual Promotion (using the Restormer backbone). Evaluated on the intermediate deblurring target ($I_{\text{tilt}}$), the joint optimization within D²Turb consistently provides beneficial geometric constraints to the texture backbone.}
\label{tab:ablation_mutual}
\centering
\begin{tabularx}{\linewidth}{@{} l *{3}{>{\centering\arraybackslash}X} @{}}
\toprule
\multirow{2}{*}{\textbf{Deblurring Module Optimization}} & \multicolumn{3}{c}{\textbf{Synthetic Dataset ($I_{\text{tilt}}$ target)}} \\
\cmidrule(l){2-4}
& PSNR$\uparrow$ & SSIM$\uparrow$ & LPIPS$\downarrow$ \\
\midrule
Standalone Deblur & 28.43 & 0.8100 & 0.2151 \\
Joint Deblur within D²Turb & \textbf{28.57} & \textbf{0.8184} & \textbf{0.1882} \\
\bottomrule
\end{tabularx}
\end{table}

Fundamentally, D²Turb's explicit disentanglement and interactive feature injection foster mutual promotion between texture and geometry recovery. To empirically prove this, we evaluate the intermediate deblurring capability.. We compare a standalone deblurring model against the texture module extracted from our fully jointly-trained D²Turb framework.

As shown in Table \ref{tab:ablation_mutual}, the in-framework module remarkably outperforms the standalone baseline across all metrics, achieving significantly better intermediate $I_{tilt}$ recovery on the synthetic dataset. This confirms that gradients flowing backward from the downstream rectification loss provide invaluable geometric constraints. It actively guides the upstream backbone to preserve edge topologies conducive to subsequent spatial unwarping, proving the universal validity of our theoretical loop.

\section{Conclusion}
We present D²Turb, a unified single-frame framework that circumvents the distortion-perception trade-off by explicitly decoupling texture recovery from geometric rectification. To bridge the sim-to-real gap, our Depth-Aware Simulation engine provides crucial ``tilt'' supervision, while the proposed ASPI mechanism dynamically aligns deep texture priors to guide precise spatial unwarping. Consequently, D²Turb establishes a robust, physically-grounded state-of-the-art across synthetic and real-world datasets. However, modeling extreme high-frequency non-rigid deformations remains challenging for our lightweight rectifier, and recovering completely obliterated structures necessitates temporal cues. Future work will explore more expressive deformation models (\eg, deformable attention~\cite{xia2022vision} or foundational flow models~\cite{teed2020raft,xu2022gmflow}) and extend this explicitly decoupled paradigm to multi-frame scenarios.

%
%
\bibliographystyle{splncs04}
\bibliography{main}
\end{document}